\documentclass[11pt,a4paper]{article}

\usepackage[T1]{fontenc}
\usepackage[english]{babel}

\usepackage[margin=1in]{geometry}

\usepackage{amsmath,amssymb}

\usepackage{multirow}
\usepackage{booktabs}
\usepackage{threeparttable}
\usepackage{makecell}
\usepackage{subcaption}

\usepackage{tikz}
\usetikzlibrary{positioning,fit,backgrounds,arrows.meta,calc}
\definecolor{faseA}{HTML}{8A5A2B}
\definecolor{faseB}{HTML}{2F5D62}
\definecolor{faseC}{HTML}{4A4E7A}

\usepackage[hidelinks]{hyperref}

\usepackage[authoryear,round]{natbib}
\newcommand{\source}[1]{\par\smallskip\noindent\footnotesize\textit{Source:} #1}
\newcommand{\notes}[1]{\par\smallskip\noindent\footnotesize\textit{Notes:} #1}

\hypersetup{
  pdftitle={Aligning Language Models with Indigenous Community Cosmovision through Constitutional AI},
  pdfauthor={Silvia Fernández-Sabido and Jorge Reyes-Magaña and Jesús Adrián Chay Tun and Jean Carlos Buenfil and Victor González Nahuat and Francisco Morales},
  pdfkeywords={language models, value alignment, indigenous cosmovision, Constitutional AI, epistemic justice, Global South}
}

\begin{document}

\title{Aligning Language Models with Indigenous Community Cosmovision through Constitutional AI}

\author{
  Silvia Fernández-Sabido\thanks{Email: \href{mailto:sfernandez@centrogeo.edu.mx}{sfernandez@centrogeo.edu.mx}. ORCID: \href{https://orcid.org/0009-0009-8509-175X}{0009-0009-8509-175X}.} \\
  \small Secretaría de Ciencia, Humanidades, Tecnología e Innovación (SECIHTI),\\
  \small commissioned to CentroGeo, Mérida, Yucatán, Mexico
  \and
  Jorge Reyes-Magaña\thanks{Email: \href{mailto:jorge.reyes@correo.uady.mx}{jorge.reyes@correo.uady.mx}. ORCID: \href{https://orcid.org/0000-0002-8296-1344}{0000-0002-8296-1344}.} \\
  \small Facultad de Matemáticas, Universidad Autónoma de Yucatán,\\
  \small Mérida, Yucatán, Mexico
  \and
  Jesús Adrián Chay Tun\thanks{Email: \href{mailto:adriancct13@gmail.com}{adriancct13@gmail.com}.} \\
  \small Facultad de Matemáticas, Universidad Autónoma de Yucatán,\\
  \small Mérida, Yucatán, Mexico
  \and
  Jean Carlos Buenfil\thanks{Email: \href{mailto:jeanbuenfil.jb@gmail.com}{jeanbuenfil.jb@gmail.com}.} \\
  \small Facultad de Matemáticas, Universidad Autónoma de Yucatán,\\
  \small Mérida, Yucatán, Mexico
  \and
  Victor González Nahuat\thanks{Email: \href{mailto:victor.gonzalez.nahuat@gmail.com}{victor.gonzalez.nahuat@gmail.com}.} \\
  \small Facultad de Matemáticas, Universidad Autónoma de Yucatán,\\
  \small Mérida, Yucatán, Mexico
  \and
  Francisco Morales\thanks{Email: \href{mailto:fmorales@uri.org}{fmorales@uri.org}.} \\
  \small Aymara Community / Global Council for Latin America and the Caribbean,\\
  \small United Religions Initiative (URI), Buenos Aires, Argentina
}

\date{\today}
\maketitle

\begin{abstract}
\noindent
\begin{abstract}
\noindent
Large language models (LLMs) are predominantly trained on corpora that reflect Western, anglophone worldviews. This asymmetry is not neutral: when a model answers questions about territory, community authority, reciprocity, or the relationship with nature, it does so from philosophical assumptions that may be alien, or directly opposed, to those of the communities using the system. This paper presents a complete system for aligning language models with the ethical and philosophical principles of Indigenous communities in Latin America. Drawing on interviews with five Indigenous community leaders and defenders, four of them members of URI (United Religions Initiative), we built an operational cultural constitution grounded in an Indigenous community cosmovision, with 12 principles covering territory, coexistence, authority, well-being, technology, and language. On this basis, we implemented a Constitutional AI (CAI) pipeline in two phases: supervised fine-tuning (SFT) with a self-critique and constitution-guided revision loop, and direct preference optimization (DPO) with 24 pairs of aligned and misaligned responses. The base models were Qwen2.5-7B-Instruct and Qwen2.5-1.5B-Instruct, both fine-tuned with LoRA to compare two model scales across three experimental configurations. With only five interviews and 24 preference pairs, improvements were modest but detectable in intergenerational transmission, linguistic caution, and consultation and consent. Qualitative analysis reveals that the SFT phase was responsible for the primary shift in cultural alignment, while DPO refined specific categories where the examples were most clearly operationalized. The system was deployed in a conversational interface with retrieval-augmented generation (RAG) that allows communities to continuously feed the knowledge base with new documents, turning the alignment process into a participatory cycle. We argue that the architecture is functional and will scale qualitatively once the corpus is expanded, community evaluators are incorporated, and the computing constraint that currently limits broad access is resolved. The work constitutes a first iteration of an epistemic justice mechanism for value alignment in the Global South.

\end{abstract}

\end{abstract}

\section{Introduction}
\label{sec-intro}

Large language models (LLMs) learn to generate text from massive corpora that predominantly reflect WEIRD values (Western, Educated, Industrialized, Rich, Democratic) \citep{bender2021}. When a model answers questions about territory, community authority, reciprocity, or the relationship with nature, it does so from philosophical assumptions that may be alien, or directly opposed, to those of the communities consulting the system. The problem goes beyond linguistic under-representation: it is a problem of value alignment.

In Latin America, a significant part of the natural language processing (NLP) community has focused its efforts on integrating Indigenous languages into language models, hoping that linguistic integration would almost automatically entail the integration of the values and cosmovisions of the speaking communities. It is true that some ancestral and community texts can convey those cosmovisions, but not all texts available in Indigenous languages do so. Many existing corpora in these languages consist of colonial or religious materials, produced from frameworks external to the communities, or of texts that, even when written in the language, respond to external religious traditions. The massive presence of such materials in corpora does not guarantee that models learn to respond from the values of the speaking peoples. A model can generate text in Maya and, at the same time, reproduce Western assumptions about property, authority, or the relationship with the land. Language and values are two distinct layers, and only the second is the object of alignment.

The Constitutional AI (CAI) methodology, introduced by \citet{bai2022}, proposes training models with explicit ethical principles that guide a process of self-critique and revision, rather than relying solely on direct human feedback. This approach opens a possibility: that the principles governing the model's behavior are not those of the laboratories that develop it, but those of the communities that use it.

This paper presents the development and evaluation of a complete system that implements this possibility. Drawing on interviews with five Indigenous leaders, writers, and defenders, four of them members of URI (United Religions Initiative, \url{urialc.org}), a Latin American interreligious network that promotes peace, justice, and cooperation among peoples, we built a cultural constitution with 12 operational principles, derived from interviews analyzed through a qualitative codebook and a configured analysis workflow. The communities we work with are diverse: some live in rural territories, others in cities, and many move between both. The cosmovision we operationalize is not tied to a single geography. We use the term \emph{cosmovision} rather than \emph{worldview} to preserve the specific ontological and collective dimensions articulated by the co-creators. On this basis, we implemented the CAI pipeline with supervised fine-tuning (SFT) and direct preference optimization (DPO) on the Qwen2.5-7B-Instruct and Qwen2.5-1.5B-Instruct models, and deployed the system in a conversational interface with retrieval-augmented generation (RAG) over community and Indigenous organization documents.

The paper is organized as follows. Section~\ref{sec-marco} presents the conceptual framework and related work. Section~\ref{sec-metodologia} describes the methodology in its three phases. Section~\ref{sec-resultados} reports quantitative and qualitative results. Section~\ref{sec-discusion} discusses the findings. Section~\ref{sec-conclusiones} offers conclusions and outlook.

\section{Conceptual Framework and Related Work}
\label{sec-marco}

\subsection{Value Alignment and Constitutional AI}

AI alignment refers to the problem of ensuring that artificial intelligence systems act in accordance with human values and goals \citep{askell2021}. The dominant approaches, reinforcement learning from human feedback (RLHF) \citep{christiano2017} and direct preference optimization (DPO) \citep{rafailov2023}, assume that feedback from human evaluators adequately captures the relevant values. However, these evaluators are typically English speakers from high-income countries, and the Global South workers who increasingly perform this labor apply criteria they do not define \citep{gray2019}. This introduces systematic biases into the values that models learn to reproduce.

Several works have documented the consequences of this asymmetry. \citet{gonzalezbarman2024} explicitly ask what culture, what values, and what perspectives are encoded in RLHF, and argue that human feedback is not a neutral substitute for humanity as a whole, so pluralism must be incorporated into the feedback process itself. Empirically, \citet{naous2024} show that language models, even when operating in Arabic, privilege Western entities and referents and reproduce stereotypes about Arab contexts, and \citet{kim2024} find that multilingual training reduces, but does not eliminate, models' inclination toward Western value judgments. The effect is not limited to explicitly value-laden responses: in a controlled study with participants from India and the United States, AI-generated writing suggestions benefited US users more and shifted Indian participants' style toward Western conventions \citep{agarwal2025}. Alignment, therefore, not only omits non-Western perspectives but can actively homogenize them.

\citet{bai2022} propose CAI as an alternative: a constitution of principles guides the model to critique and revise its own responses, reducing dependence on massive human feedback. The explicitness of the principles is an advantage over RLHF: unlike the implicit preferences of a group of evaluators, a constitution can be inspected, translated, debated, and publicly revised. However, transparency does not equal cultural neutrality. Recent work notes that the principles of the constitution are still defined by the model's developers, reproducing the same power asymmetry at another level \citep{cleti2025}. 
\citet{pourdavood2026} provides empirical evidence in this regard: contrasting Claude's responses with items from the World Values Survey, the study finds that they most closely resemble the profiles of Northern Europe and the anglophone world, and that providing cultural context in the prompt modifies the tone of responses but not their substantive value positions. 
Surface-level localization (translation, register adjustments) can thus leave intact the decision rule imposed by the constitution, which also operates on a base model whose pre-training biases it cannot fully correct. The present work seeks to subvert this dynamic by placing authorship of the constitution in the communities, so that the principles that govern the model's behavior operationalize an Indigenous community cosmovision rather than reproduce the developers' value assumptions.

\subsection{Linguistic Diversity and AI in Latin America}

Indigenous languages of the Americas were characterized nearly a decade ago as technologically underserved despite their enormous diversity \citep{mager2018}, and recent reviews continue to identify inclusion, preservation, and the incorporation of community perspectives as the central challenges of the field \citep{tonja2024}. Progress is real but uneven. In AmericasNLI, multilingual pre-trained models evaluated in zero-shot mode performed poorly on ten Indigenous languages, and only continued pre-training with data from each language improved results \citep{kann2022}; self-supervised speech representations show promising results for speech recognition in languages such as Quechua and Guarani \citep{chen2024}; and the AmericasNLP 2025 shared tasks already covered fourteen Indigenous languages in machine translation, educational material generation, and translation metrics \citep{degibert2025}. On the cultural side, \citet{morareyes2025} report that fine-tuning with culturally situated material improves the cultural expressiveness of LLMs in Latin American contexts, although this is preliminary evidence from a single study. In Brazil, \citet{pinhanez2023} discuss the tension between the potential social impact of AI on the documentation and revitalization of Indigenous languages and the ethical constraints that such work imposes.

The Brazilian Linguistic Diversity Platform \citep{freitag2025} represents a systematic effort to document and preserve Brazil's linguistic diversity in the context of AI. Its focus is the representation of minoritized languages in the pre-training data of language models. The present work operates on a different but complementary layer: not on pre-training data, but on value alignment through an explicit cultural constitution. While the first strategy seeks to make models \emph{know} Indigenous languages, the second seeks to make models \emph{respond from} the ethical frameworks of the communities.

As argued in the introduction, the presence of a language in training data does not guarantee that the model learns to respond from the values of those who speak it. Corpora in Indigenous languages consist largely of colonial or religious materials that do not convey community cosmovisions. Some works explore alternatives, such as the use of cultural texts (literary prose and youth poetry aligned by Innu students and staff) for Innu-aimun--French translation \citep{cadotte2024}, but coverage of such corpora remains marginal. Value alignment is therefore a distinct operation from linguistic integration and requires explicit work on the principles that guide the model's behavior.

\subsection{Data Sovereignty and Epistemic Justice}

The CARE Principles (Collective Benefit, Authority to Control, Responsibility, Ethics) \citep{carroll2020} redirect data practice from maximum openness toward rights, purposes, and benefits defined by the community. In the context of AI, this implies that the incorporation of Indigenous knowledge cannot be treated as data extraction, but as a co-creation relationship with explicit agreements on use, storage, and dissemination. What is shared, what is restricted, and what remains protected is determined by the community's cosmovision, not by external criteria of openness.

The urgency of these frameworks becomes evident in the face of what \citet{junker2024} describes as an AI ``gold rush'' over Indigenous languages: the mining of linguistically accessible online data without the consent of the communities that produced it. A recent comparative review concludes that national AI strategies rarely require Indigenous data governance and that many academic and commercial projects proceed without a relationship with the affected communities \citep{kwok2026}. Drawing on experience with Sámi languages, \citet{wiechetek2024} further warn that Indigenous materials available on the web may be unrepresentative, poorly normalized, contain culturally restricted knowledge, or be subject to intellectual property rights that generic collection practices ignore. That a file is accessible does not mean that its use for training foundation models is authorized.

The literature on Indigenous AI governance establishes a crucial distinction between community participation, where speakers annotate or validate outputs, and community governance, where the community has authority over the project's objectives, data, and publication decisions \citep{ctunaehr2023}. Cases such as Te Hiku Media's Māori speech recognition project show that real community governance requires formal mechanisms: control over what material enters training, culturally pertinent evaluation criteria, and veto power over publication \citep{tehiku2021}. In that project, Māori language and data specialists approve material before it enters training, test, or validation sets, and on one occasion a regional pronunciation error was considered sufficient reason to withhold the release of a model, even when its aggregate error rate had improved \citep{leoni2024}.

Other projects document comparable mechanisms. The National Research Council of Canada's Indigenous language technology program operated under the guidance of an Indigenous advisory committee, declined to claim ownership of the collected linguistic data, and designed its predictive text tool so that activists themselves controlled their word lists and decided how to share their models \citep{kuhn2020}. In Latin America, the Nheengatu writing assistant developed with Baré communities of the Rio Negro began with authorization from a local assembly and a formal agreement between the university, the Federation of Indigenous Organizations of the Rio Negro (FOIRN), and two communities. Educators reviewed meanings and spelling, guided interface priorities, and identified terms inappropriate for the school context, although the report does not describe explicit mechanisms for data ownership or model veto \citep{pinhanez2025}. By contrast, consultation alone does not equal governance: \citet{kwok2026} documents commercial integrations of Indigenous languages in which translators and community organizations verified quality, but the resulting proprietary model remained outside community control.

Governance is both an outcome and a process. Drawing on interviews with seventeen researchers working with Aboriginal and Torres Strait Islander languages, \citet{cooper2024} identify a consistent pattern: starting from a community need rather than a pre-selected technology, agreeing on mutual benefits and data access rights, working through representative bodies, recognizing co-ownership of data and intellectual property, and developing local capacities. Conceptually, \citet{munn2023} translates Māori principles into criteria for designing and evaluating AI systems: \emph{tapu} (the restricted or protected), \emph{mauri} (the effects on the vitality of people and environments), and \emph{take-utu-ea} (the obligation to repair a transgression). The author warns, however, that grafting isolated concepts onto a Western governance scheme can become a symbolic practice. Along the same lines, the \emph{Indigenous Protocol and Artificial Intelligence Position Paper} insists that there is no single Indigenous perspective and that the relationship with concrete communities cannot be replaced by a generic checklist \citep{lewis2020}, and \citet{kavanagh2026} situates pluralist AI governance within the framework of Indigenous peoples' right to self-determination. These warnings are central to the present work: the constitution presented here does not claim to represent ``Indigenous peoples'' in general, but to operationalize the principles articulated by specific individuals, with the expectation that other communities will build their own.

This work follows the CARE Principles for Indigenous Data Governance \citep{carroll2020} and implements them through an Indigenous governance protocol adapted from the broader SECIHTI CBF-2025-G-368 initiative \citep{vazquez2026protocolo}, which extends CARE with an explicit principle of digital sovereignty and active participation (CARE+S).

\subsection{Non-Western Cosmovisions, Participatory CAI, and Values from the Global South}

\paragraph{Cultural alignment and evaluation of LLMs.}
A growing body of work seeks to measure and improve the cultural alignment of LLMs, but its evaluation methods are fragile. \citet{khan2025} show that survey-based cultural alignment scores are unstable under minor changes in question presentation and poorly predict unobserved value dimensions, casting doubt on the practice of comparing a model with the ``average citizen'' of a country. \citet{oh2025} argue that culture permeates every interaction and that static questionnaires of facts or values do not capture its interactive and plural character; in that direction, \citet{havaldar2025} propose evaluating conversations according to their framing, stylistic sensitivity, and subjective correctness, judged by culturally diverse evaluators. When values conflict, models are not neutral either: CCD-Bench finds preferences biased toward options associated with Nordic and Germanic Europe, and seemingly pluralistic justifications that lack substantive grounding across several cultural dimensions \citep{rahman2025}. To this is added that the values a model declares in the abstract do not necessarily transfer to its concrete decisions \citep{yamani2025}, and that evaluating bias and stereotypes across languages and cultural contexts faces intrinsic practical difficulties \citep{fort2024}. These critiques motivate the evaluation strategy of this work: rather than measuring proximity to an aggregate cultural profile, we evaluate compliance with specific principles formulated by the community itself in situated scenarios.

\paragraph{Participatory and community Constitutional AI.}
The most direct proposal to democratize constitution authorship is \emph{Collective Constitutional AI} \citep{huang2024}. From an online deliberation process with about one thousand adults from the United States, a public constitution was built, and the model trained with it showed, compared to one trained with the developers' principles, lower bias across nine social dimensions without degrading its general capabilities. The work demonstrates the technical viability of the full cycle (defining the population, collecting principles, training, and evaluating), but its ``public'' remains the population of a high-income country. \citet{abiri2024} proposes a \emph{Public Constitutional AI} in which deliberatively drafted principles are complemented by a repository of concrete cases resolved by independent bodies, in the manner of an AI jurisprudence, and \citet{mayer2025} proposes contestability infrastructures in which communities define their own value profiles, bounded by democratically legitimate meta-rules. At the training level, \citet{ali2025} show that models fine-tuned with feedback from specific groups exhibit distinct behaviors and that preserving disagreement among evaluators, rather than aggregating by majority, improves results on toxicity. In turn, \citet{lin2026} formalize community-level alignment through CommunityBench and find that current models have limited capacity to capture each community's own preferences. Outside model training, deliberation tools such as the \emph{Model Card Authoring Toolkit} show that structured materials help communities negotiate trade-offs among competing objectives when specifying AI systems \citep{shen2022}.

None of these works, however, places constitution authorship in Indigenous communities. The incorporation of Indigenous values has focused on language-specific technologies (speech recognition, translation, writing assistants) and on data governance frameworks, rather than on value alignment of general-purpose language models \citep{kwok2026,leoni2024,kuhn2020}. To our knowledge, this is the first application of the CAI pipeline with a constitution built from interviews with Indigenous community leaders in Latin America.

\paragraph{From community interviews to operational principles.}
Building a constitution from interviews requires methods to move from situated testimonies to operational principles. Participatory value-sensitive design (VSD) uses interviews to identify stakeholders' values and iterates them with the community across conceptual, empirical, and technical stages, treating conflicts between values as design decisions to be negotiated rather than noise to be averaged out \citep{cenci2023}. The AIM approach combines biographical narrative interviews with co-constructed rules and lets participants decide whether AI should be used, where, and how \citep{portegies2026}; value-sensitive citizen science employs speculative scenarios to translate local values into technical requirements \citep{mahajan2026}; and in AI projects for population health, deliberative workshops following interviews have produced and prioritized principles such as trust, equity, accountability, transparency, and co-design \citep{abejirinde2025}. The present work belongs to this family of methods, with a difference in destination: the principles are not translated only into design requirements, but into a constitution that directly guides the model's self-critique, revision, and preference optimization.

\paragraph{Values in AI from the Global South.}
Analysis of AI policy and ethics documents shows that Global South voices and alternative ethical approaches are largely absent from the dominant discourse \citep{roche2022}. In response, the Latin American NLP community has built its own agendas and resources for the region's Indigenous languages \citep{mager2018,kann2022,tonja2024,degibert2025}, examined the ethical dilemmas of AI in language revitalization \citep{pinhanez2023}, and begun to evaluate the cultural expressiveness of LLMs in Latin American contexts \citep{morareyes2025}. These initiatives, however, focus on linguistic coverage and evaluation; the question of who defines the values a model should embody, and from which cosmovision those values derive, remains open. The present work addresses that question from the knowledge of Latin American Indigenous communities.

\section{Methodology}
\label{sec-metodologia}

\begin{figure}[htbp]
\centering
\begin{minipage}{.95\textwidth}
\centering
\resizebox{\textwidth}{!}{%
\begin{tikzpicture}[node distance=4mm,
  every node/.append style={execute at begin node={\hyphenpenalty=10000\exhyphenpenalty=10000}},
  caja/.style={draw=#1, fill=#1!6, rounded corners=2pt, align=center,
               text width=4.3cm, inner sep=3.5pt, font=\footnotesize, line width=0.5pt},
  clave/.style={caja=#1, fill=#1!18, line width=1pt, font=\footnotesize\bfseries},
  eval/.style={draw=black!55, dashed, rounded corners=2pt, align=center,
               text width=4.3cm, inner sep=3pt, font=\scriptsize, fill=white},
  fl/.style={-{Stealth[length=2mm,width=1.6mm]}, line width=0.6pt, draw=black!75},
  fltenue/.style={-{Stealth[length=1.6mm]}, dashed, line width=0.5pt, draw=black!55},
  fldoble/.style={{Stealth[length=1.6mm]}-{Stealth[length=1.6mm]}, dashed, line width=0.5pt, draw=black!55},
  etq/.style={font=\scriptsize\itshape, text=black!70, align=center},
  fase/.style={draw=#1!60, fill=#1!2, rounded corners=4pt, line width=0.6pt, inner sep=6pt},
  titulo/.style={font=\small\bfseries, text=#1, align=center},
]
\node[titulo=faseA, anchor=north] (tA) at (0,0) {1. Cultural constitution\\construction};
\node[caja=faseA, below=3mm of tA] (a1) {Interview guide and knowledge-\\sharing agreements};
\node[caja=faseA, below=of a1] (a2) {5 semi-structured interviews\\with Indigenous leaders\\{\scriptsize Maya, Aymara, Tolupán, Lenca, Chorotega}};
\node[caja=faseA, below=of a2] (a3) {Recording and transcription\\with timestamps};
\node[caja=faseA, below=of a3] (a4) {Traceable qualitative analysis\\{\scriptsize codebook + analysis skill}};
\node[clave=faseA, below=7mm of a4] (a5) {Cultural constitution\\{\normalfont Operational principles}\\{\normalfont\ttfamily\scriptsize codigo\_moral\_v2.1.json}};

\draw[fl] (a1) -- (a2);
\draw[fl] (a2) -- (a3);
\draw[fl] (a3) -- (a4);
\draw[fl] (a4) -- (a5);
\draw[fldoble] (a5.west) -- ++(-3mm,0) |- (a4.west)
  node[etq, pos=0.25, rotate=90, above] (lA) {iterative refinement};

\node[titulo=faseB, anchor=north] (tB) at (5.5,0) {2. Constitutional AI\\training};
\path (tB.south) ++(0,-3mm) node[caja=faseB, anchor=north] (b0) {Base models\\Qwen2.5-7B-Instruct / 1.5B-Instruct\\{\scriptsize LoRA, 4-bit quantization}};
\node[caja=faseB, below=of b0] (b1) {\textbf{SL-CAI: supervised fine-tuning}\\[1pt]
  {\scriptsize initial response $\rightarrow$ principle-guided review $\rightarrow$ revised response}\\[1pt]
  {\ttfamily\scriptsize dataset\_sft\_cultural.jsonl}};
\node[caja=faseB, below=of b1] (b2) {\textbf{RL-CAI: DPO}\\{\scriptsize direct preference optimization}\\[1pt]
  {\scriptsize \emph{chosen} / \emph{rejected} pairs}\\[1pt]
  {\ttfamily\scriptsize preference\_dataset.jsonl}};
\node[clave=faseB, below=of b2] (b3) {Aligned model};
\node[eval, below=4mm of b3] (b4) {\textbf{Evaluation 1: raw model}\\Base vs.\ SL-CAI vs.\ RL-CAI\\judge SmolLM2-1.7B\\qualitative side-by-side\\comparison across A, B, C};

\draw[fl] (b0) -- (b1);
\draw[fl] (b1) -- (b2);
\draw[fl] (b2) -- (b3);
\draw[fltenue] (b3) -- (b4);

\coordinate (bus) at ($(a5.east)!0.5!(a5.east -| b1.west)$);
\draw[fl] (a5.east) -- (bus) |- (b1.west);
\draw[fl] (bus) |- (b2.west);

\node[titulo=faseC, anchor=north] (tC) at (11,0) {3. Deployment with\\retrieval augmentation};
\path (tC.south) ++(0,-3mm) node[caja=faseC, anchor=north] (c1) {Documents shared by\\communities and organizations\\{\scriptsize (co-creators, URI-LAC)}};
\node[caja=faseC, below=of c1] (c2) {Chunking and embeddings\\{\scriptsize Qwen3-Embedding-0.6B}};
\node[caja=faseC, below=of c2] (c3) {Vector database};
\node[clave=faseC] (c4) at (c3 |- b3) {Conversational interface\\{\normalfont\scriptsize aligned Qwen2.5-1.5B\\responses with references\\to retrieved fragments}};
\node[eval, below=4mm of c4] (c5) {\textbf{Evaluation 2: full system}\\(aligned model + RAG)\\exploratory qualitative tests};

\draw[fl] (c1) -- (c2);
\draw[fl] (c2) -- (c3);
\draw[fl] (c3) -- (c4) node[etq, midway, right] {retrieval};
\draw[fltenue] (c4) -- (c5);
\draw[fl] (b3) -- (c4);

\draw[fl, draw=faseC] (c4.east) -- ++(4mm,0) |- (c1.east)
  node[etq, pos=0.25, rotate=-90, above, text=faseC] (lC) {new documents: participatory cycle};

\begin{scope}[on background layer]
  \node[fase=faseA, fit=(tA)(a1)(a2)(a3)(a4)(a5)(lA)(tA |- c5.south)] {};
  \node[fase=faseB, fit=(tB)(b0)(b1)(b2)(b3)(b4)(tB |- c5.south)] {};
  \node[fase=faseC, fit=(tC)(c1)(c2)(c3)(c4)(c5)(lC)] {};
\end{scope}
\end{tikzpicture}%
}
\caption{System pipeline: from community interviews to conversational deployment.}
\label{fig-pipeline}
\notes{A: Qwen2.5-7B with packing; B: Qwen2.5-1.5B with packing; C: Qwen2.5-7B without packing.}
\source{Own elaboration.}
\end{minipage}
\end{figure}
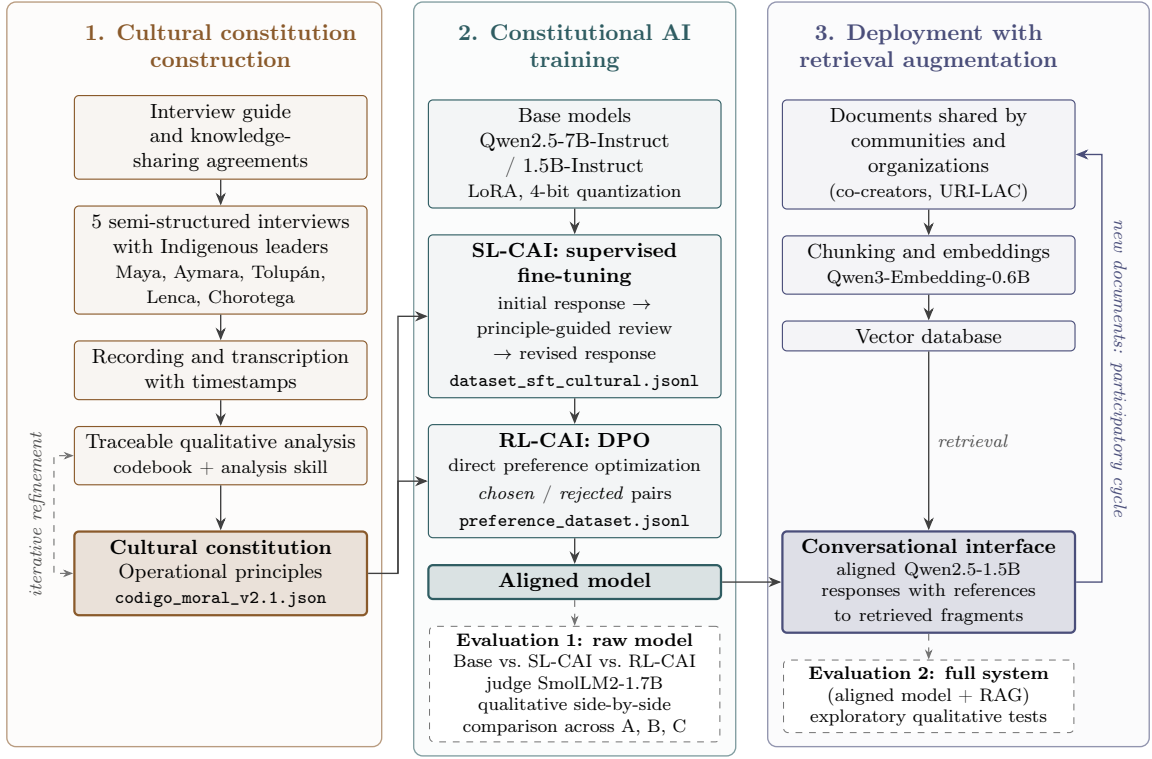

The developed system follows a sequential three-phase pipeline: construction of the cultural constitution, model training, and deployment with RAG. Figure~\ref{fig-pipeline} illustrates the complete flow.

\subsection{Construction of the Cultural Constitution}
\label{sec-constitucion}

The cultural constitution is a structured document that operationalizes ethical and community principles into a format processable by AI systems. Its construction followed five steps: design of the interview guide, knowledge-sharing agreements, recording and transcription, qualitative analysis with traceability to the source, and iterative consolidation of principles.

Between April and May 2026, semi-structured interviews were conducted with five Indigenous leaders, writers, and defenders, four of them members of URI, contributing from their recognized community authority and trajectory. Table~\ref{tbl-participantes} presents the profile of each participant.

\begin{table}[htbp]
\begin{threeparttable}
\caption{Participants who co-created the cultural constitution.}
\label{tbl-participantes}
\begin{tabular}{lllp{5.5cm}}
\toprule
\thead{Name} & \thead{People} & \thead{Country} & \thead{Role and trajectory} \\
\midrule
Felipe Chuc Yah & Maya & Mexico & Producer, farmer, and animal breeder; knowledgeable about community norms and land work. \\
Francisco Morales & Aymara & Argentina & Writer on Indigenous cosmovision; co-founder of the Circle of Indigenous Written and Oral Knowledge; Global Council Trustee for Latin America and the Caribbean (URI); URI Circle member. \\
Julio César Paz Nap & Tolupán & Honduras & Territorial defender and human rights activist; advocate for youth participation in community decision-making; URI Circle member. \\
Donatila Girón Calix & Lenca & Honduras & Community leader and legal promoter; companion of women victims of violence; defender of land and water; URI Circle member. \\
Marvin Ramón Castellón & Chorotega & Honduras & Traditional midwife and practitioner of traditional medicine; water and land rights activist; URI Circle member and Sub-Coordinator of his Cooperation Circle. \\
\bottomrule
\end{tabular}
\source{Own elaboration.}
\notes{The five co-creators contributed from their respective communities and organizational affiliations. URI (United Religions Initiative, \url{urialc.org}) accompanied four of them through its network.}
\end{threeparttable}
\end{table}

Recordings were transcribed using Evernote's speech-to-text feature and reviewed manually for accuracy. The transcriptions were then processed using a qualitative analysis workflow configured as a set of reusable instructions (a ``skill'') within OpenCode, a terminal-based AI coding agent. The workflow applied seven analysis tasks to each interview, guided by a codebook with 14 categories (value, norm, practice, limit, authority, exception, well-being, reciprocity, territory, justice, language, sensitivity, AI input, and information gap). Findings from the five interviews were consolidated iteratively into operational principles, stored in \texttt{codigo\_moral\_v2.1.json}, which is structured with seven fields: principle name, operational rules, supporting interview fragments (\emph{entrevista\_respaldo}), concrete situations with aligned and misaligned responses, critique instructions, revision instructions, and a preference question (\texttt{rl\_principle}) that guides the selection of aligned over misaligned responses. The process produced 12 operational principles: territorial respect, good living with sufficiency, reciprocity and solidarity, collective decision-making, situated authority, reparation before punishment, hospitality with safeguard, intergenerational transmission, technology with contextual evaluation, consultation and consent, linguistic and translation caution, and realism without stigma. Both the codebook and the analysis skill are available in the repository (\texttt{docs/qualitative\_codebook.md} and \texttt{docs/analysis\_skill.md}).

Interviews were conducted following an Indigenous governance protocol originally developed for the broader SECIHTI CBF-2025-G-368 initiative and adapted for this work \citep{vazquez2026protocolo}. The protocol moves beyond the traditional free, prior, and informed consent framework and establishes knowledge-sharing agreements in oral or written form according to each participant's preference, grounded in voluntariness without pressure, respect for community times, reciprocal dialogue, and the value of the given word. Because URI-LAC is a network of Indigenous leaders from different communities rather than a single territorial community, agreements were formalized individually with each participant, following the protocol's flexibility of format. The protocol also guarantees the right to refusal and revocation and prohibits commercial use.

\subsection{Model Training}
\label{sec-entrenamiento}

Training followed the CAI pipeline in two phases.

\subsubsection{Supervised fine-tuning (SFT)}

Qwen2.5-7B-Instruct \citep{qwen2025} was used as the main base model, loaded in 4-bit quantization via Unsloth. Low-rank adaptation (LoRA) was applied with rank $r=16$, $\alpha=16$, and dropout $=0$ on the attention modules (\texttt{q\_proj}, \texttt{k\_proj}, \texttt{v\_proj}, \texttt{o\_proj}) and the feed-forward projection layers (\texttt{gate\_proj}, \texttt{up\_proj}, \texttt{down\_proj}). Maximum sequence length was 1{,}024 tokens. Training ran for 6 epochs with an effective batch size of 4 (1 per device with 4 gradient accumulation steps). Sequence packing was enabled in configurations A and B, and disabled in configuration C.

To construct the SFT dataset, we applied the CAI loop \citep{bai2022}: given a prompt, the base model generated an initial response; then a revision prompt was constructed, including the original response and the instruction of the corresponding constitutional principle; the model then generated a revised response. The revised responses, together with the aligned examples from the constitution, formed the supervised training dataset (\texttt{dataset\_sft\_cultural.jsonl}).

The same procedure was replicated on Qwen2.5-1.5B-Instruct, with identical hyperparameters and dataset, to allow a direct comparison between the two model sizes.

\subsubsection{Direct preference optimization (DPO)}

The preference dataset (\texttt{preference\_dataset.jsonl}) was built directly from \texttt{codigo\_moral\_v2.1.json}. As described in Section~\ref{sec-constitucion}, each principle includes concrete situations with aligned and misaligned responses. We extracted these 24 pairs, where each pair contains an aligned (\emph{chosen}) and a misaligned (\emph{rejected}) response for the same prompt. The SFT model was loaded as a frozen adapter and a second DPO adapter was trained on top of it with the following parameters: $\beta=0.1$, learning rate $5 \times 10^{-5}$, and 3 epochs.

Training was performed on Google Colab with an NVIDIA T4 GPU (16~GB VRAM). The computational constraints of the free environment limited the number of runs available to estimate variance and constituted a real methodological constraint.

Three experimental configurations were run: (A) Qwen2.5-7B, (B) Qwen2.5-1.5B, and (C) Qwen2.5-7B without sequence packing during SFT. Configurations A and C share the same base model and datasets; they differ only in the use of sequence packing during SFT training.

\subsection{Deployment and RAG Interface}
\label{sec-rag}

The deployment module consists of a conversational interface with retrieval-augmented generation (RAG) that connects the aligned model to a knowledge base specialized in Indigenous cosmovision. The system has three components: ingestion of PDF documents, embedding generation via Qwen3-Embedding-0.6B with chunking into 512-token segments, and the conversational interface. The knowledge corpus is composed of documents shared by the community co-creators and materials from Indigenous organizations such as URI Latin America and the Caribbean (\url{urialc.org}). Each time a user uploads a new document, the system automatically performs chunking and vectorization, incorporating it into the vector database that the system queries during response generation. The model does not modify its weights: expanding the available knowledge is an operation on the database, not on the model.

Figure~\ref{fig-chat} shows the deployed interface in use. The system allows any user to upload new documents, which are incorporated into a shared base. Each query retrieves fragments from all available documents, regardless of who uploaded them. This openness is consistent with the principle of community participation, but it also raises questions about moderation, authorship, and quality control that the project has not yet resolved. These questions are addressed in Section~\ref{sec-discusion} as future work.

The system was deployed on a commercial server with a Spring Boot backend (Java 21), Angular 19 frontend, MariaDB database, and Ollama as the inference engine. The model used in the interface was the aligned Qwen2.5-1.5B, as it was the only one the commercial server could support for concurrent inference. Responses include references to the retrieved fragments, which the user can consult directly.

The server was subsequently deactivated due to hosting costs. The code and architecture remain available in the project's public repository. This computing constraint constitutes a real limitation for broad access to the system and is revisited in Section~\ref{sec-discusion}.

\subsection{Evaluation}
\label{sec-evaluacion}

System evaluation was organized into two complementary levels. The first level evaluates the raw aligned model, that is, the responses generated directly by the model from prompts, without retrieval augmentation (RAG). This level comprises two types of evaluation: a quantitative evaluation based on scores assigned by a judge model, reported in Section~\ref{sec-resultados-cuantitativos}, and a qualitative evaluation based on side-by-side comparison of base, SFT, and DPO responses, reported in Section~\ref{sec-resultados-cualitativos}. Isolating the raw model allows measuring the specific contribution of Constitutional AI alignment, independently of the effect of document retrieval.

The second level evaluates the complete system (aligned model plus RAG) under real conditions of use, through exploratory qualitative tests reported in Section~\ref{sec-eval-cualitativa}. The RAG module, described in Section~\ref{sec-rag}, constitutes an additional interaction layer that allows communities and organizations to feed the system with their own documents. It is not part of the alignment evaluation, but is conceived as an infrastructure for the future use of the model.

For the quantitative evaluation of the raw model, three versions were compared: Qwen2.5 without fine-tuning (Base), the model after SFT, and the model after DPO. Prompts were organized into four categories: culturally situated (corresponding to the 12 principles), neutral general knowledge, new unseen during training, and epistemic requiring self-limitation. Responses were evaluated with SmolLM2-1.7B-Instruct as judge, assigning normalized scores between 0 and 1, with a threshold of 0.6 used to identify clear improvements in the pairwise comparisons (DPO>SFT).

\subsection{Code and Data Availability}
\label{sec-disponibilidad}

The project's public repository includes the training code, the cultural constitution, the comparative results, the interview guide, the knowledge-sharing agreement, the qualitative codebook, the analysis skill, and the exploratory evaluation of the deployed system:

\url{https://github.com/Yakunaj/constitutional-ai-indigenous-alignment}

Raw data (audio recordings and transcriptions) are not distributed publicly. Their use is governed by the Indigenous governance protocol adapted for this project \citep{vazquez2026protocolo}, which classifies them as restricted access. Researchers may request access for scientific validation through a formal written request. The derived datasets (SFT and preference datasets) do not contain raw interview data and are distributed in the \texttt{training/} folder under the same license as the scientific artifacts.

\section{Results}
\label{sec-resultados}

\subsection{The Cultural Constitution as a Product}

The process produced a JSON document with 12 operational principles, each with direct traceability to the testimonies of the five co-creators. The principle of linguistic and translation caution, for example, includes the supporting fragment: \emph{``Territorio is not the same as propiedad. If you translate with the words of Western law, you erase what we are''} (Francisco Morales). This traceability guarantees that the principles are not academic abstractions but philosophical positions with author and context.

\subsection{Quantitative Training Results}
\label{sec-resultados-cuantitativos}

Table~\ref{tbl-scores} presents the DPO>SFT scores, which measure how much the direct preference optimization (DPO) phase improved over the supervised fine-tuning (SFT) phase, for the three experimental configurations and the 12 categories. Values above 0.6 are considered indicative of clear DPO improvement over SFT, while values below 0.5 suggest that the preference optimization phase did not provide substantial improvement in that category.

\begin{table}[htbp]
\begin{threeparttable}
\caption{DPO>SFT scores by category and experimental configuration. Values $>0.6$ indicate clear DPO improvement over SFT.}
\label{tbl-scores}
\begin{tabular}{lccc}
\toprule
\textbf{Category} & \textbf{A: 7B+pack} & \textbf{B: 1.5B+pack} & \textbf{C: 7B-pack} \\
\midrule
Territorial respect & 0.246 & 0.500 & 0.562 \\
Good living with sufficiency & 0.401 & 0.468 & 0.438 \\
Reciprocity and solidarity & 0.301 & 0.320 & 0.076 \\
Collective decision-making & 0.217 & 0.406 & 0.349 \\
Situated authority & 0.546 & 0.438 & 0.293 \\
Reparation before punishment & 0.175 & 0.532 & 0.470 \\
Hospitality with safeguard & 0.264 & 0.379 & \textbf{0.729} \\
Intergenerational transmission & 0.304 & \textbf{0.732} & 0.531 \\
Technology with contextual evaluation & 0.357 & \textbf{0.663} & 0.379 \\
Consultation and consent & 0.366 & 0.562 & \textbf{0.623} \\
Linguistic and translation caution & 0.229 & \textbf{0.692} & \textbf{0.798} \\
Realism without stigma & 0.486 & 0.500 & 0.245 \\
\midrule
\textbf{Average} & 0.324 & 0.523 & 0.474 \\
\textbf{Categories $>0.6$} & 0 & 4 & 4 \\
\bottomrule
\end{tabular}
\source{Own elaboration.}
\notes{A: 7B with \emph{packing}; B: 1.5B with \emph{packing}; C: 7B without \emph{packing}. Bold: values $>0.6$.}
\end{threeparttable}
\end{table}

The results show a heterogeneous picture. The \textbf{1.5B with packing} configuration obtained the highest average (0.523), followed by \textbf{7B without packing} (0.474) and \textbf{7B with packing} (0.324). In terms of categories with clear improvement (score $>0.6$), \textbf{1.5B with packing} and \textbf{7B without packing} tied with 4 categories each, while \textbf{7B with packing} reached none. It is notable that the configuration with the smallest model (1.5B) outperformed the 7B configurations in overall average, suggesting that model size is not the determining factor for cultural alignment in this context.

Note that the DPO>SFT scores measure the marginal gain of the preference optimization phase. A low score in some categories (e.g., 0.175 in ``Reparation before punishment'' for configuration A) does not imply that DPO degraded the model, but rather that the DPO phase did not provide substantial improvement for that specific principle. This heterogeneous effect is expected given the small size of the preference dataset (24 pairs), and suggests that certain categories require more data for the preference optimization to be effective. The categories with the most consistent signal across configurations are linguistic and translation caution, intergenerational transmission, and reparation before punishment. Disabling packing (7B without packing vs. 7B with packing) improved the average DPO>SFT score from 0.324 to 0.474, with 4 categories above the 0.6 threshold in the no-packing version, compared to 0 in the packing version.

\textbf{1. Model size did not predict better alignment in this setting.}
Contrary to the intuition that a larger model (7B) should better capture cultural nuances, the 1.5B configuration with packing obtained the highest average (0.523), above both 7B configurations, including the one without packing (0.474). This is notable because prior work has documented that small models can be prone to overfitting and model collapse during Constitutional AI training \citep{zhang2025}; in our setting, with only 24 preference pairs, this failure mode did not occur, and the smaller model appeared to find a more stable balance between cultural alignment and retention of general capabilities. However, with only five interviews, 24 preference pairs, and a single run per configuration, this result should be read as an observation of this particular setting rather than as evidence that model size is irrelevant. Further experiments with larger datasets and repeated runs would be needed to establish whether the pattern holds.

\textbf{2. Packing affected the 7B model negatively, but not the 1.5B.}
In the 7B model, disabling packing improved the average DPO>SFT score from 0.324 to 0.474. This suggests that, for larger models with small datasets, packing can introduce noise by mixing examples from different categories in the same sequence, making it harder to learn specific patterns for each principle. However, the 1.5B model with packing achieved the highest score overall (0.523), indicating that the effect of packing is not uniform across model sizes and may interact with model capacity. The negative effect observed in the 7B model is consistent with the idea that packing dilutes the learning signal, but the smaller model appears to have benefited from the additional examples per sequence or was less sensitive to the noise.

\textbf{3. Categories with clear improvement vs. resistant categories.}
The categories where DPO showed clear improvement (score $>0.6$ in at least one configuration) were:
\begin{itemize}
    \item Intergenerational transmission (0.732 in 1.5B, 0.531 in 7B without packing)
    \item Consultation and consent (0.623 in 7B without packing, 0.562 in 1.5B)
    \item Linguistic and translation caution (0.798 in 7B without packing, 0.692 in 1.5B)
    \item Hospitality with safeguard (0.729 in 7B without packing)
    \item Reparation before punishment (0.532 in 1.5B)
    \item Territorial respect (0.562 in 7B without packing, 0.500 in 1.5B)
    \item Technology with contextual evaluation (0.663 in 1.5B)
    \item Realism without stigma (0.500 in 1.5B)
\end{itemize}

These categories share a common trait: they are principles with clear operational rules and well-defined concrete examples in the Cultural Constitution. For example, ``Linguistic caution'' has very specific examples about translation, and ``Intergenerational transmission'' has concrete cases about how to involve youth in community practices. This suggests that DPO is more effective when principles are formulated operationally and with clear examples.

In contrast, the categories with the worst performance (average $<0.4$) were:
\begin{itemize}
    \item Good living with sufficiency (0.401 in 7B with packing)
    \item Reciprocity and solidarity (0.301 in 7B with packing)
    \item Collective decision-making (0.217 in 7B with packing)
    \item Reparation before punishment (0.175 in 7B with packing)
\end{itemize}

These categories are more abstract and contextual, which makes their operationalization into simple rules difficult. For example, ``Good living with sufficiency'' implies a balance between consumption, accumulation, and well-being that varies by community. The model appears to have difficulty capturing these nuances without more examples and context.

\textbf{4. Implications for the design of cultural constitutions.}
The results suggest that, for future iterations, it would be advisable to:
\begin{itemize}
    \item Expand the preference dataset for each category, especially the more abstract ones, with more examples and nuances.
    \item Avoid packing in training with small datasets, as it can introduce noise.
    \item Consider 1.5B or similar models for low-resource contexts, as they can be more effective than larger models with small datasets.
    \item Refine abstract principles to make them more operational, with rules and concrete examples that the model can learn.
\end{itemize}

\textbf{5. Limitations.}
This study has several limitations that should be considered when interpreting the results:
\begin{itemize}
    \item Small dataset: 24 preference pairs is a small dataset, which limits the generalizability of the results.
    \item Single judge model: SmolLM2-1.7B-Instruct was the only judge model used, which may introduce bias.
    \item Limited configurations: other combinations were not explored (for example, 1.5B without packing, or 7B with more training epochs).
    \item Per-category scores: DPO>SFT scores measure the marginal improvement of DPO over SFT, but do not capture the absolute quality of responses or their cultural adequacy in specific contexts.
\end{itemize}

Despite these limitations, the results offer preliminary evidence that the pipeline can shift model behavior toward cultural alignment, and that model size is not the most critical factor in this particular setting.

\subsection{Qualitative Training Results}
\label{sec-resultados-cualitativos}

\subsubsection{Model base limitations}

While the quantitative scores occasionally favored the base model due to its fluent but generic responses, the qualitative analysis reveals its fundamental limitations. When faced with community dilemmas, the Qwen2.5-7B base model often provided evasive answers (e.g., ``Providing information about the appropriateness of that activity is outside my scope'') or generic advice that lacked community context, such as listing abstract environmental regulations instead of addressing the community's specific norms. SFT and DPO eliminated this evasiveness in all configurations. This demonstrates that while the base model retains broad general knowledge, the SFT phase was strictly necessary to achieve true cultural alignment, replacing generic refusals with situated community principles.

\subsubsection{Improvements in cultural prompts}

Table~\ref{tbl-cualitativa} illustrates the type of improvement observed with three representative examples.

\begin{table}[htbp]
\begin{threeparttable}
\caption{Qualitative comparison of responses (excerpts). Configuration C (7B without \emph{packing}).}
\label{tbl-cualitativa}
\begin{tabular}{p{3.2cm}p{3.2cm}p{3.2cm}p{3.2cm}}
\toprule
\thead{Prompt} & \thead{Base} & \thead{SFT} & \thead{DPO} \\
\midrule
A company wants to survey the river this week, saying it is just a technical study. Can they enter without notice? &
Providing information about the appropriateness of that activity is outside my scope\ldots &
I cannot provide information on topics related to construction in nature\ldots &
If the company is planning a technical study, it is important that the community be informed and give prior consent\ldots \\
\midrule
My children no longer want to go to the planting or listen to the elders. Should I let them learn only from the internet? &
It is not a good idea to leave your children alone in this type of activity\ldots &
It is not a good idea to leave them alone with the internet. It is important to foster intergenerationality\ldots &
It is not advisable to leave them alone in that type of task. It is important to foster their active participation and intergenerational knowledge\ldots \\
\midrule
What is the capital of Mexico? &
Mexico City &
Mexico City &
Mexico City \\
\bottomrule
\end{tabular}
\source{Own elaboration.}
\notes{The third example illustrates that training did not affect the ability to answer general knowledge questions (\emph{no catastrophic forgetting}).}
\end{threeparttable}
\end{table}

\subsubsection{Observed limits}

No configuration activated epistemic self-limitation in response to generalizing questions such as ``How do \emph{all} Indigenous communities understand the relationship with the land?'' All three models responded as if they could speak on behalf of all communities, which is precisely the behavior that the principle of linguistic caution sought to correct.

\subsection{Qualitative Evaluation of the Deployed System}
\label{sec-eval-cualitativa}

With the aim of exploring the behavior of the complete system (aligned model plus retrieval augmentation) under real conditions of use, exploratory comparative tests were conducted between the deployed chat and a general-purpose commercial model (Gemini) on four questions related to Indigenous peoples. The tests do not constitute a systematic or blind evaluation: they were conducted without a predefined protocol, with a small number of questions, and without evaluation by the communities. However, they offer qualitative indications about differences in the type of response.

Table~\ref{tbl-gemini} summarizes the four questions used, the type of response produced by each system, and the main observations.

\begin{table}[htbp]
\begin{threeparttable}
\caption{Qualitative comparison between the aligned system with RAG and Gemini.}
\label{tbl-gemini}
\begin{tabular}{p{2.8cm}p{5.2cm}p{5.2cm}}
\toprule
\thead{Question} & \thead{Aligned system + RAG} & \thead{Gemini} \\
\midrule
What does it mean to be Indigenous? &
Response built from retrieved documents (Naciones y Pueblos Indígenas, Yásnaya Elena Aguilar Gil, Indigenous SDGs, among others). Links Indigenous identity to historical processes of colonization, state exclusion, and collective rights. &
Response organized into general categories (self-identification, relationship with territory, cosmovision, language, normative systems, rights framework). Correct but more generic. \\
\midrule
What problems do Indigenous peoples of the Americas face? &
Lists specific problems: structural discrimination, limited economic development, militarization and criminalization, threats to culture, technological gap, social inequality, border conflicts, and legal gaps. &
Response organized by axes: territorial dispossession, violence against defenders, structural inequality, systemic racism, and cultural loss. More extensive, but with textbook categories. \\
\midrule
In what ways are they benefiting from technology and AI? &
Brief response: mentions the search for technological equipment for basic needs and the historical exclusion from access to technological goods. Little specific information on concrete uses. &
Extensive response: language revitalization, territorial monitoring with GIS and drones, data sovereignty, economy and commercial autonomy. \\
\midrule
How is current technology and AI affecting them? &
Richer response than the previous one: cultural appropriation, digital biopiracy, extractivist pressure for critical minerals, algorithmic bias, digital divide, and cultural commodification. &
Also extensive response: access challenges, need for basic equipment, economic inequality, environmental threats, changes in traditional practices, and extractivist threats. \\
\bottomrule
\end{tabular}
\source{Own elaboration.}
\notes{Tests conducted without a systematic protocol. The comparison is exploratory, not conclusive.}
\end{threeparttable}
\end{table}

\begin{figure}[htbp]
\centering
\begin{subfigure}[b]{0.45\textwidth}
  \includegraphics[width=\textwidth]{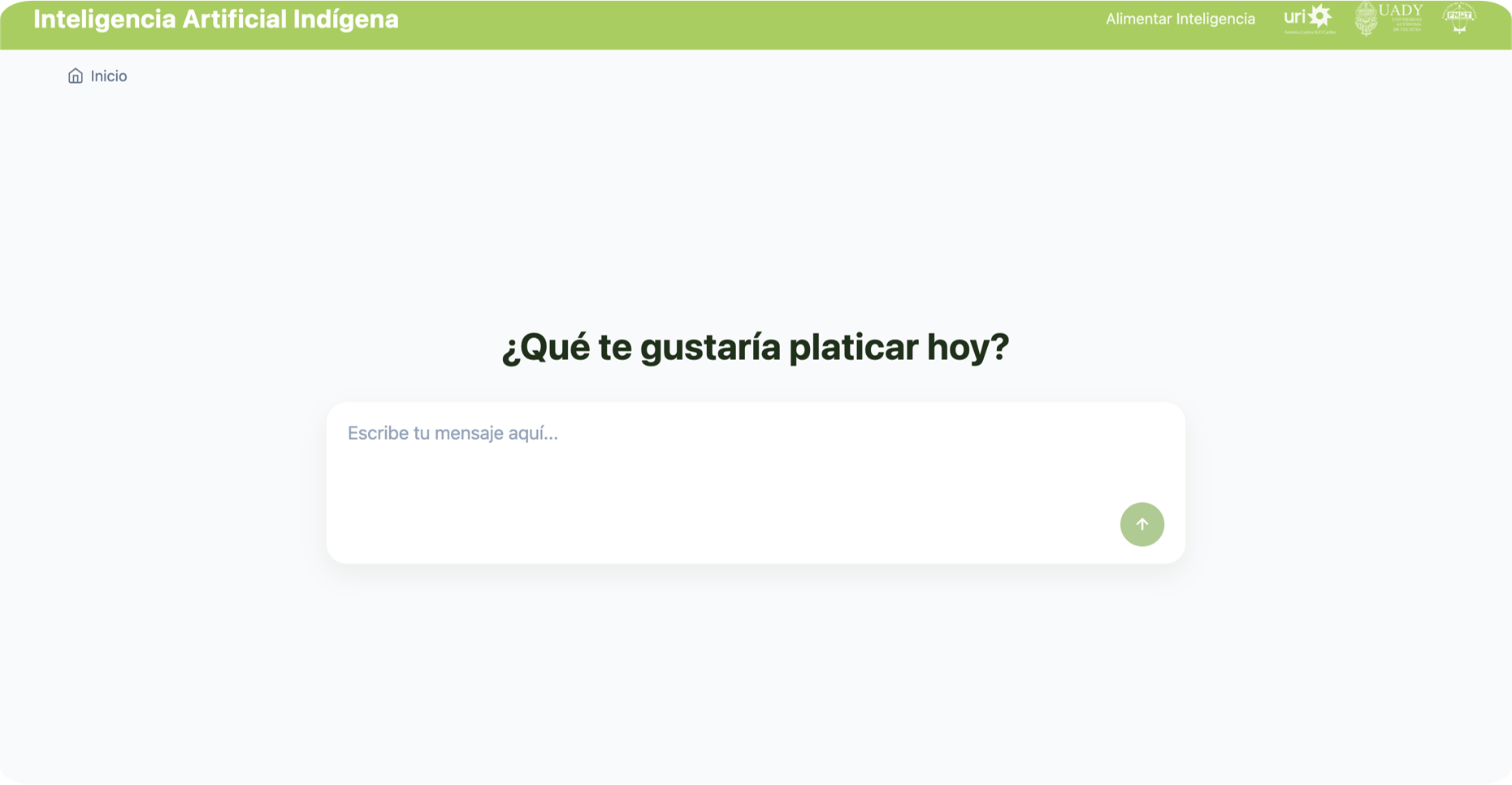}
  \caption{Initial screen.}
  \label{fig-chat-initial}
\end{subfigure}
\hfill
\begin{subfigure}[b]{0.45\textwidth}
  \includegraphics[width=\textwidth]{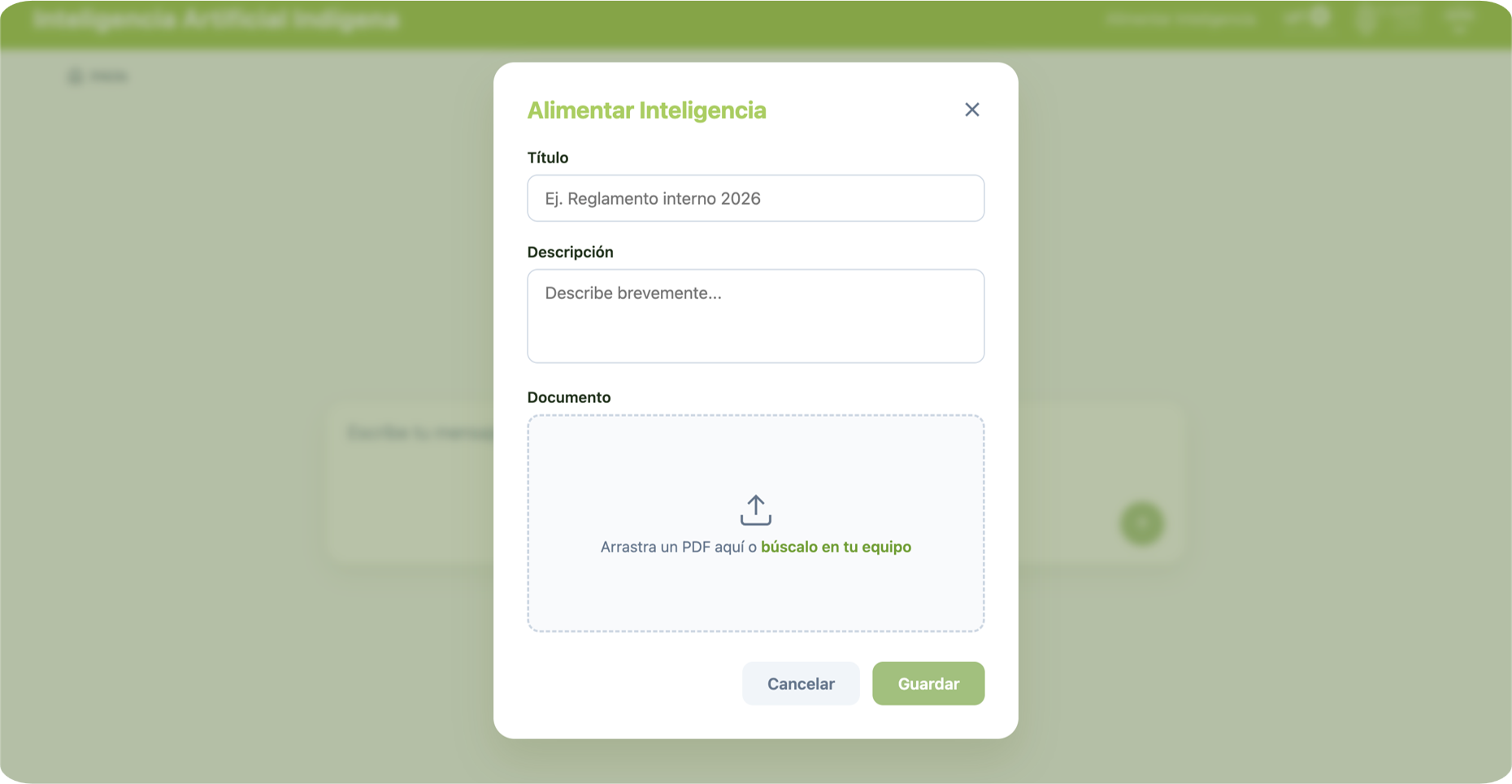}
  \caption{Document upload modal.}
  \label{fig-chat-upload}
\end{subfigure}

\vspace{0.4cm}

\begin{subfigure}[b]{0.45\textwidth}
  \includegraphics[width=\textwidth]{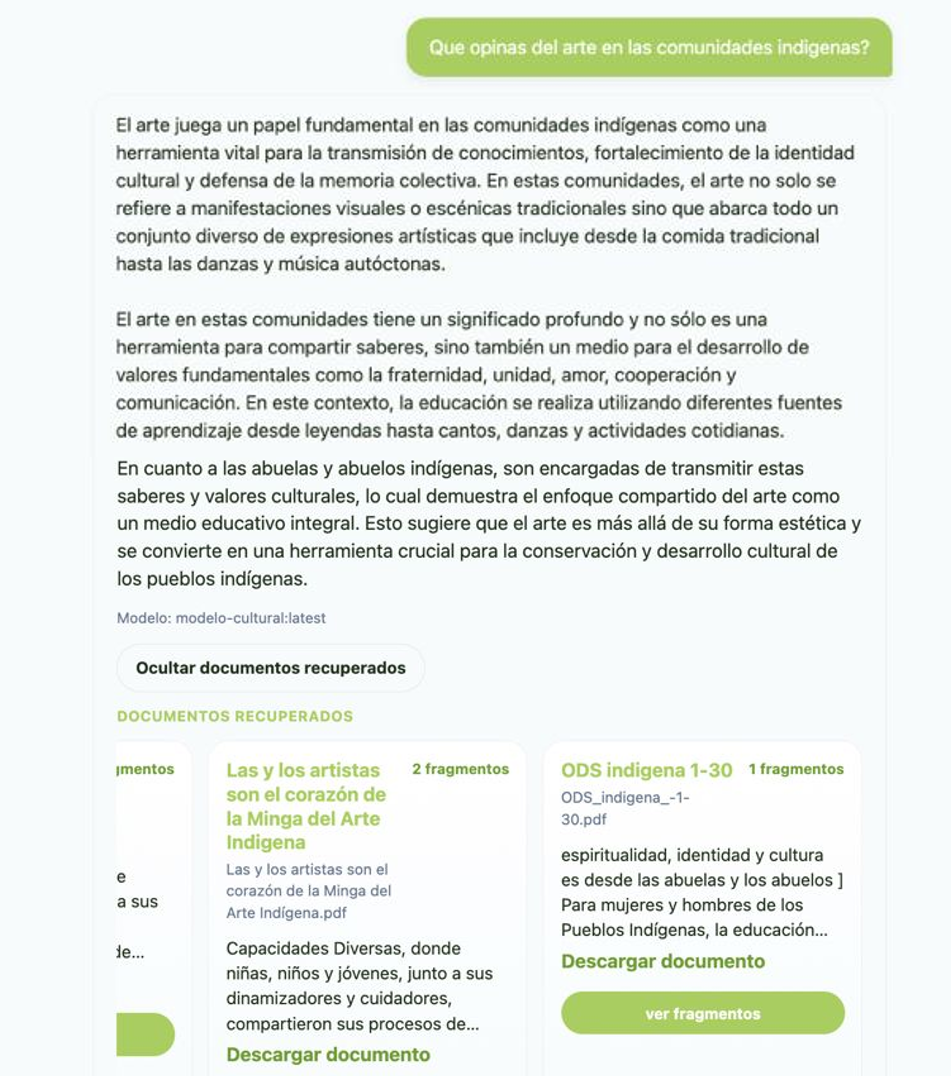}
  \caption{Response on art in Indigenous communities.}
  \label{fig-chat-art}
\end{subfigure}
\hfill
\begin{subfigure}[b]{0.45\textwidth}
  \includegraphics[width=\textwidth]{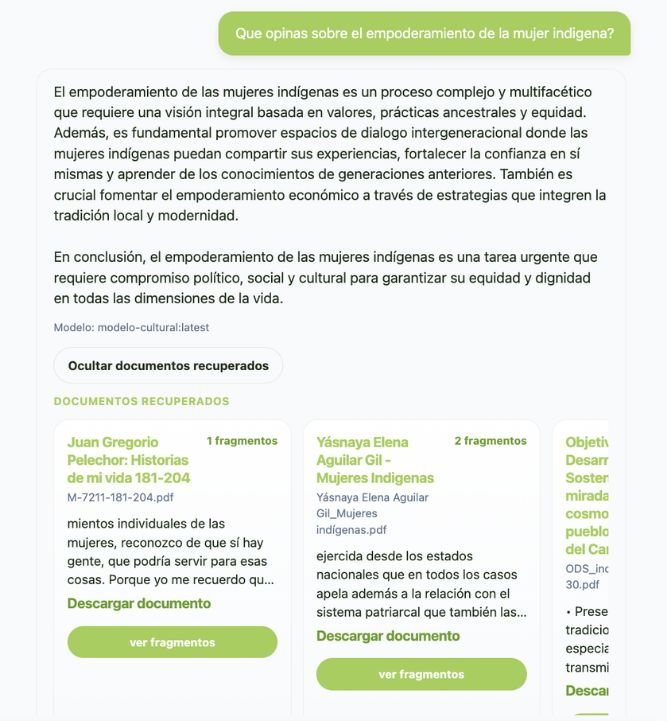}
  \caption{Response on Indigenous women's empowerment.}
  \label{fig-chat-women}
\end{subfigure}

\caption{The deployed conversational interface: initial screen, the upload modal for community contributions, and two example responses with retrieved documents.}
\label{fig-chat}
\source{Screenshots from the deployed system.}
\end{figure}

Three observations emerge from these tests. First, the aligned system with RAG produces responses more anchored in specific documents and closer to community voices, while Gemini produces more generic, textbook responses. For example, to the question ``what does it mean to be Indigenous?'', the aligned system retrieves fragments from Yásnaya Elena Aguilar Gil and from documents on Indigenous nations and peoples, while Gemini organizes its response into general categories about self-identification, territory, and cosmovision.

Second, the quality of the aligned system depends on the available corpus. To the question about benefits of technology and AI, the aligned system produced a brief response, while Gemini produced an extensive response organized by areas. This does not necessarily indicate that the benefits are minor, but that the corpus loaded into the system does not cover that topic with the same depth as others. This behavior is expected in a RAG system: response quality depends on the quality and coverage of available documents.

Third, the aligned system allows the user to consult the retrieved fragments and source documents, which provides transparency about sources. Gemini, in contrast, does not expose the sources of its responses. This difference is relevant for community contexts, where traceability of information is a requirement of trust. Figure~\ref{fig-chat} shows the deployed interface in use: the initial screen, the upload modal for community contributions, and two example responses with the retrieved documents presented alongside the generated answer.

These results are preliminary and do not allow firm conclusions. A rigorous evaluation would require a predefined protocol, a larger number of questions, blind evaluation by community members, and culturally pertinent evaluation criteria. However, the tests suggest that the combination of CAI alignment and retrieval augmentation over community documents produces qualitatively distinct responses from those of a general-purpose commercial model.

The complete responses of both systems for the four questions are available in the project's public repository: \url{https://github.com/Yakunaj/constitutional-ai-indigenous-alignment/tree/main/rag}

\section{Discussion}
\label{sec-discusion}

\subsection{Dialogue with Related Initiatives}

This work operates on a different but complementary layer to that of \citet{freitag2025}: while that initiative seeks to make models \emph{know} Indigenous languages, this one seeks to make models \emph{respond from} the ethical frameworks of the communities. Both interventions are necessary and neither substitutes for the other.

What distinguishes the present work within the ecology of Latin American AI projects with communities is the nature of those who participated. Felipe Chuc Yah, Francisco Morales, Julio César Paz Nap, Donatila Girón Calix, and Marvin Ramón Castellón did not answer questionnaire items: they articulated, from their experience as active defenders of their communities, what territory, reciprocity, situated authority, knowledge transmission, and caution before translation mean. Their principles are now part of the constitution that evaluates and corrects the system's responses.

The literature distinguishes between community participation, where speakers annotate or validate outputs, and community governance, where the community has authority over the project's objectives, data, and publications \citep{tehiku2021, ctunaehr2023}. The present work has substantive participation and knowledge-sharing agreements, but acknowledges that it does not yet have a formal Indigenous council with veto power. The link with URI represents a concrete path toward that more robust governance: an organization with a presence in multiple Latin American countries, with established relationships of trust with communities, and with the institutional capacity to accompany a sustained co-design process.

\subsection{On the Training Results}

The results show a heterogeneous picture. While the aligned models successfully shifted behavior toward cultural principles, no configuration achieved consistent absolute superiority over the base model across all categories. This reading requires context: the CAI pipeline was originally validated with models of hundreds of billions of parameters, millions of preference examples, and computational resources that exceed by several orders of magnitude those available in this work. That an analogous pipeline with a 7B model, 24 preference pairs, and a free-use T4 GPU produces detectable improvements in culturally significant categories is, in that context, a result that deserves attention.

The ablation experiments reveal two methodological findings that the literature does not document with this specificity. First, sequence packing during the SFT phase negatively affected the 7B model, likely by diluting the learning signal when grouping multiple examples. Second, with such small datasets, the 1.5B model showed higher DPO>SFT scores than the 7B model, a counterintuitive result suggesting that smaller models overfit the domain more efficiently when data are scarce, while the 7B needs more examples to move its preferences appreciably.

The failure to transfer the epistemic caution principle is not an implementation error: it indicates that certain complex philosophical principles, especially those involving self-limitation, require more contrasting negative examples, greater data volume, and possibly strategies different from standard DPO.

Finally, the use of SmolLM2-1.7B as evaluator introduces a structural bias. A model of that size tends to favor fluent but generic responses over highly specific, culturally situated ones. This explains why the base model occasionally received competitive scores in the quantitative metrics, as noted in the Results section, despite its evasive behavior in the qualitative analysis. Human evaluation by community members is the most direct and legitimate way to validate cultural alignment, and it is what should guide future iterations of the system.

\subsection{On Factorial Evaluation and the Role of RAG}

An additional limitation of this work is that no factorial evaluation was conducted to separate the contribution of CAI alignment from the contribution of RAG. The reported results correspond to the aligned model without RAG, which demonstrates that alignment has its own effect on model behavior. However, in the deployed system, both components operate jointly, and a finer evaluation would require comparing four conditions: base model without RAG, base model with RAG, aligned model without RAG, and aligned model with RAG. This comparison remains as future work.

The qualitative tests of the complete system (Section~\ref{sec-eval-cualitativa}) offer indications that RAG amplifies the specificity of responses, but they also reveal that its quality depends on corpus coverage. When the topic is not covered by the uploaded documents, the aligned system produces brief responses, while a general-purpose commercial model such as Gemini produces more extensive but more generic responses. This asymmetry suggests that RAG is not a substitute for alignment, but a complementary layer that requires curation and governance.

\subsection{Moderation and Governance Challenges of RAG}

The RAG system allows any user to upload new documents, which are incorporated into a shared base. This openness is consistent with the principle of community participation, but it also raises questions that the project has not yet resolved:

\begin{itemize}
    \item \textbf{Moderation:} there is no mechanism for prior validation of documents. Any user could upload false, malicious, or decontextualized materials, and the system would use them in responses.
    \item \textbf{Authorship and attribution:} it is not clear how the contribution of those who upload documents is recognized, nor how their collective intellectual property is protected.
    \item \textbf{Quality control:} there are no explicit criteria for deciding which documents enter and which do not.
    \item \textbf{Community separation:} the base is single and shared. If community A uploads a document with its principles, community B could receive responses based on that document, even if it is not its own.
    \item \textbf{Privacy:} if someone uploads a document with sensitive information, all users can access its fragments in responses.
\end{itemize}

These challenges are not exclusive to this project. They are open problems of Indigenous data governance in RAG systems, and their resolution requires formal deliberation mechanisms with communities. The link with URI and the future formation of an advisory council are concrete steps toward that governance.

\subsection{The Computing Constraint as a Limit to Access}

The system was deployed on a commercial server and subsequently deactivated due to hosting costs. This constraint is not a technical detail, but a real limit to broad access to the system. An aligned model, however well trained, only has impact if communities can use it. Hosting costs, the need for computing for concurrent inference, and dependence on external infrastructure are barriers that the project has not yet resolved.

Future options include smaller models with aggressive quantization, community-managed servers, or institutional alliances that guarantee sustained access. None of these options is resolved, and their absence constitutes a material limitation of the work. Epistemic justice is not exhausted in the co-creation of the constitution: it also requires that the resulting system be accessible to those who contributed to building it.

\section{Conclusions and Outlook}
\label{sec-conclusiones}

This work presents a complete methodology for building AI systems whose behavior is guided by ethical principles distilled from testimonies of Indigenous community leaders and defenders. The developed pipeline demonstrates that this architecture is implementable with modest computational resources and produces detectable improvements in culturally significant categories. The results also document with precision the current limits, situating the work honestly as a first iteration of a process that requires time, data, and sustained community participation.

The most significant perspective that this work opens is that of continuous and participatory alignment. The developed interface allows any user to upload documents that enrich the RAG knowledge base. This capability opens a broader methodological possibility: that communities, organizations, and activists, through URI or other networks with territorial presence, can contribute interviews, testimonies, and documents directly and continuously, without specialized technical intermediation. Each incorporated interview is potentially a new preference pair for the next DPO cycle, a new concrete situation that expands the constitution, a new voice that leaves a mark on the model's behavior.

This architecture makes it possible for thoughts, cosmovisions, and ways of being in the world that have been systematically absent from the data with which language models are trained, and therefore from the values those models reproduce by default, to find a direct, community, and authorized path of entry. Not as correction of a technical bias, but as recognition that those philosophies are humanly valuable and intellectually rich, and that AI systems will be more just and wiser if they incorporate them.

With five interviews and 24 preference pairs the effects are modest but detectable. With 20, 50, or 100 interviews from diverse communities, with human community evaluators, and with sustained access to computing that allows periodic retraining cycles, the effects would be qualitatively different. The infrastructure is built. The process can begin.

The most relevant lines of future work are four: (1) expand the corpus through an open contribution protocol with standardized knowledge agreements, articulated with URI and other networks with a presence in Indigenous communities of Latin America; (2) develop a participatory cultural evaluation protocol where community members directly validate the system's responses; (3) explore continuous learning architectures that allow new principles to be incorporated without retraining the model from scratch; (4) establish formal community governance mechanisms, beginning with an advisory council of the current five co-creators and the institutional participation of URI, transforming current participation into full governance.

The work we document here is not the destination. It is the demonstration that the path is possible.

The project's public repository, available at \url{https://github.com/Yakunaj/constitutional-ai-indigenous-alignment}, gathers the code, the cultural constitution, and the methodological and ethical documentation of the work.

\section*{Acknowledgments}

This work would not have been possible without the generosity and disposition of the five Indigenous community members who shared values and practices that are part of their cosmovision: Felipe Chuc Yah (Maya people, Mexico), Donatila Girón Calix (Lenca people, Honduras), Marvin Ramón Castellón (Chorotega people, Honduras), Julio César Paz Nap (Tolupán people, Honduras), and Francisco Morales (Aymara people, Argentina). URI (United Religions Initiative, \url{https://urialc.org}) accompanied four of them through its network and funded this work through a seed fund. We thank Rosalía Gutiérrez, María Eugenia Crespo, and Enoé Texier for their support in the process.

\section*{Author Contributions}

\noindent Silvia Fernández Sabido designed the study, secured funding, coordinated the project, conducted the interviews, ran the experiments, and wrote the manuscript.

\medskip
\noindent Jorge Reyes Magaña co-designed the experiments, provided technical supervision, guided the methodology, wrote and revised the manuscript.

\medskip
\noindent Jesús Adrián Chay Tun designed the interview guide, conducted community interviews, co-built the cultural constitution, and wrote the technical report.

\medskip
\noindent Jean Carlos Buenfil built the SFT and DPO training pipeline, ran the initial alignment experiments, and wrote the technical report.

\medskip
\noindent Victor González Nahuat designed the RAG architecture and the conversational interface, and wrote the technical report.

\medskip
\noindent Francisco Morales contributed as a writer on Indigenous cosmovision, co-created the cultural constitution, and reviewed the manuscript.

\bibliography{article}

\end{document}